\documentclass[letterpaper, 10 pt, conference]{ieeeconf}

\IEEEoverridecommandlockouts
\usepackage{amsmath}
\usepackage{graphicx}
\usepackage{amssymb}
\usepackage{algorithm}
\usepackage{algpseudocode}
\usepackage{multirow}
\usepackage{booktabs}
\usepackage{cite}
\usepackage{url}
\usepackage{pifont} 
\usepackage{xcolor} 
\usepackage{hyperref}

\usepackage{tabularx}
\usepackage{booktabs}

\title{\LARGE \bf
Surfel-LIO: Fast LiDAR-Inertial Odometry with Pre-computed Surfels and Hierarchical Z-order Voxel Hashing
}

\author{Seungwon Choi$^{1}$, Dong-Gyu Park$^{2}$, Seo-Yeon Hwang$^{2}$ and Tae-Wan Kim$^{1}$
\thanks{$^{1}$Seungwon Choi and Tae-Wan Kim are with the Department of Naval Architecture and Ocean Engineering, Seoul National University, Seoul, Korea.
        {\tt\small \{csw3575, taewan\}@snu.ac.kr}}%
\thanks{$^{2}$Donggyu Park and Seo-Yeon Hwang are with Clobot Co., Ltd., Seoul, Korea.
        {\tt\small \{derrick, ian\}@clobot.co.kr}}%
}

\begin{document}

\maketitle
\thispagestyle{empty}
\pagestyle{empty}

\begin{abstract}

LiDAR-inertial odometry (LIO) is an active research area, as it enables accurate real-time state estimation in GPS-denied environments. Recent advances in map data structures and spatial indexing have significantly improved the efficiency of LIO systems. Nevertheless, we observe that two aspects may still leave room for improvement: (1) nearest neighbor search often requires examining multiple spatial units to gather sufficient points for plane fitting, and (2) plane parameters are typically recomputed at every iteration despite unchanged map geometry. Motivated by these observations, we propose Surfel-LIO, which employs a hierarchical voxel structure (hVox) with pre-computed surfel representation. This design enables O(1) correspondence retrieval without runtime neighbor enumeration or plane fitting, combined with Z-order curve encoding for cache-friendly spatial indexing. Experimental results on the M3DGR dataset demonstrate that our method achieves significantly faster processing speed compared to recent state-of-the-art methods while maintaining comparable state estimation accuracy. Our implementation is publicly available at \url{https://github.com/93won/lidar_inertial_odometry}.
\end{abstract}

\section{Introduction}
\label{sec:introduction}

LiDAR-Inertial Odometry (LIO) is a widely used technique for real-time state estimation in GPS-denied environments~\cite{cadena2016past, placed2023survey}. By tightly coupling high-frequency IMU measurements with precise LiDAR point clouds, LIO systems provide accurate and robust performance even under challenging conditions such as fast motion and geometrically degenerate scenes~\cite{shan2020lio, qin2020lins}.

The computational pipeline of modern LIO systems consists of three stages: IMU propagation, LiDAR-based state update, and map update. The IMU propagation step is computationally lightweight, as it involves only the integration of inertial measurements. In contrast, the LiDAR-based state update and map update stages account for the majority of the computational cost. During LiDAR-based state update, each iteration requires nearest neighbor search and plane fitting for every input point. The map update stage must also insert new points while maintaining the map structure for subsequent queries. Recent state-of-the-art methods have made remarkable progress in addressing these challenges. Fast-LIO2~\cite{xu2022fast} established a highly influential framework by employing an incremental k-d tree (ikd-Tree)~\cite{cai2021ikd} to maintain a global map, performing point-to-plane ICP~\cite{besl1992method, chen1992object, low2004linear} within an Iterated Extended Kalman Filter (IEKF)~\cite{bell1993iterated}. Unlike conventional k-d trees that require costly reconstruction when points are added, the ikd-Tree supports efficient incremental updates while maintaining $O(\log N)$ search complexity. Faster-LIO~\cite{bai2022faster} further accelerated the pipeline by replacing the tree structure with iVox (incremental Voxels), a hash-based voxel map inspired by spatial hashing techniques~\cite{niessner2013real, teschner2003optimized}, achieving $O(1)$ map insertion and demonstrating impressive real-time performance.

\begin{figure}[t]
\centering
\includegraphics[width=\columnwidth]{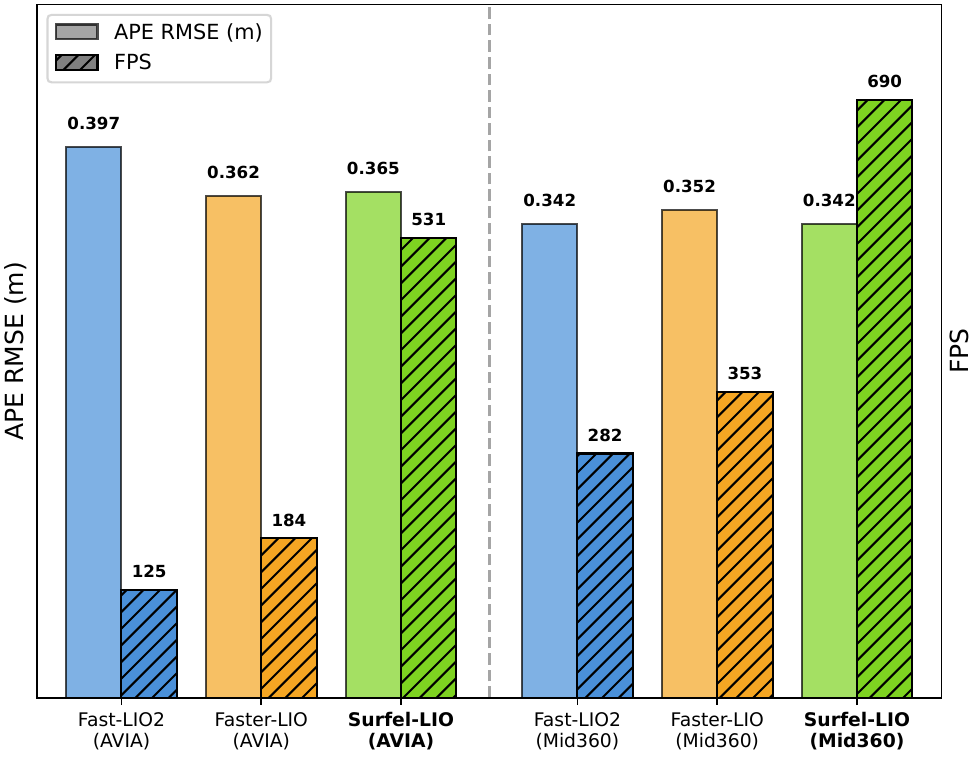}
\caption{Benchmark comparison on M3DGR dataset~\cite{zhang2025m3dgr}. Our method achieves comparable accuracy with significantly higher throughput compared to Fast-LIO2 and Faster-LIO across different LiDAR sensors.}
\label{fig:benchmark}
\end{figure}

Although these methods have achieved remarkable performance, there is still room for improvement when considering the iterative nature of IEKF-based state estimation. First, while iVox provides efficient hash-based voxel access, gathering sufficient neighbor points for plane fitting requires examining multiple surrounding voxels---typically 7 to 27 depending on configuration---which can become a bottleneck when processing dense point clouds. Second, since IEKF performs multiple iterations per scan, both methods repeatedly compute plane parameters via least-squares fitting for the same set of correspondences, incurring $O(k^2)$ overhead at every iteration where $k$ denotes the number of neighbors.

\begin{table*}[t]
\centering
\caption{Comparison of LiDAR-Inertial Odometry Methods}
\label{tab:lio_comparison}
\renewcommand{\arraystretch}{1.2}
\setlength{\tabcolsep}{5pt}
\begin{tabular*}{\textwidth}{@{\extracolsep{\fill}}>{\centering\arraybackslash}p{3.5cm}>{\centering\arraybackslash}p{4.5cm}>{\centering\arraybackslash}p{2.5cm}ccc@{}}
\toprule
\toprule
\textbf{Method} & \textbf{Estimation} & \textbf{Map Structure} & \textbf{NN Search} & \textbf{Plane Estimation} & \textbf{Map Update} \\
\midrule
LIO-Mapping \cite{ye2019tightly} & Sliding Window Optimization & k-d tree (rebuild) & $O(\log N)$ & $O(k^2)$ & $O(N \log N)$ \\
LIO-SAM \cite{shan2020lio} & iSAM2 Factor Graph & k-d tree (rebuild) & $O(\log N)$ & $O(k^2)$ & $O(N \log N)$ \\
\midrule
LINS \cite{qin2020lins} & Iterated Extended Kalman Filter & k-d tree (rebuild) & $O(\log N)$ & $O(k^2)$ & $O(N \log N)$ \\
Fast-LIO \cite{xu2021fast} & Iterated Extended Kalman Filter & k-d tree (rebuild) & $O(\log N)$ & $O(k^2)$ & $O(N \log N)$ \\
Fast-LIO2 \cite{xu2022fast} & Iterated Extended Kalman Filter & ikd-Tree & $O(\log N)$ & $O(k^2)$ & $O(\log N)$ \\
Faster-LIO \cite{bai2022faster}$^\dagger$ & Iterated Extended Kalman Filter & iVox (hash) & $O(m)$ & $O(k^2)$ & $O(1)$ \\
\midrule
\textbf{Surfel-LIO (Ours)} & \textbf{Iterated Extended Kalman Filter} & \textbf{hVox + Surfel} & $\boldsymbol{O(1)}$ & $\boldsymbol{O(1)}$ & $\boldsymbol{O(1)}$ \\
\bottomrule
\bottomrule
\end{tabular*}
\vspace{2mm}
\begin{flushleft}
\footnotesize{$N$: number of map points, $k$: number of neighbors for plane fitting, $m$: number of neighboring voxels searched (typically 7--27).\\
$^\dagger$iVox achieves $O(1)$ hash-based voxel access, but gathering sufficient neighbors requires examining multiple surrounding voxels.}
\end{flushleft}
\end{table*}

Motivated by these observations, we propose \textit{hVox}, a two-level hierarchical voxel structure built upon the efficient hash-based design of prior works. Our key idea is to pre-compute surfel parameters (centroid, normal, and covariance) at map update time, so that during IEKF iterations the system can retrieve surface geometry directly without neighbor enumeration or least-squares fitting. To ensure robust surfel estimation, we organize voxels into two levels: fine-resolution voxels (L0) alone often contain too few points for reliable normal estimation, especially in sparse regions. By grouping multiple L0 voxels into coarser L1 voxels, we accumulate sufficient points to compute stable surfel parameters while preserving fine-grained spatial resolution for accurate point localization~\cite{pfister2000surfels, hornung2013octomap, vespa2018efficient}.

For efficient spatial indexing, we employ the Z-order curve (Morton code)~\cite{morton1966computer} to encode 3D voxel coordinates into a single integer key. By leveraging the fact that consecutive LiDAR points are spatially adjacent due to the sensor's scanning pattern, the Z-order curve's locality-preserving property ensures that nearby points map to nearby hash keys, improving cache efficiency during batch queries~\cite{samet2006foundations}. Furthermore, the hierarchical relationship between L0 and L1 voxels can be computed through simple bit-shift operations, enabling efficient parent-child lookups without additional data structures.

To validate our approach, we evaluate on the M3DGR dataset~\cite{zhang2025m3dgr}, which contains sequences recorded with two distinct Livox LiDAR sensors---AVIA (non-repetitive scanning, 70° FoV) and Mid-360 (360° FoV, higher point density)---along with ground-truth trajectories from motion capture and RTK-GPS. As shown in Fig.~\ref{fig:benchmark}, by combining pre-computed surfels with cache-friendly Z-order indexing, our method runs significantly faster than both Fast-LIO2 and Faster-LIO while maintaining competitive accuracy.

In summary, the main contributions of this paper are:
\begin{itemize}
    \item A two-level hierarchical voxel map (\textit{hVox}) that pre-computes surfel parameters during map updates, enabling $O(1)$ correspondence retrieval without neighbor search or plane fitting.
    \item Z-order curve encoding that exploits LiDAR scanning locality for cache-friendly voxel access.
    \item Experimental validation on the M3DGR dataset, demonstrating improved efficiency over Fast-LIO2 and Faster-LIO while maintaining comparable accuracy.
\end{itemize}
In addition, we publicly release our implementation under a permissive license to facilitate reproducibility and encourage further research.


\section{Related Work}\label{sec:related}

LiDAR-based simultaneous localization and mapping (SLAM) research has advanced rapidly since Zhang and Singh proposed LOAM \cite{zhang2014loam}. LOAM extracts edge and planar features from 3D LiDAR point clouds based on curvature and estimates 6-DoF pose through scan-to-map matching with a k-d tree. By separating odometry and mapping into different execution frequencies, LOAM achieved both computational efficiency and accuracy. This feature extraction and k-d tree-based correspondence search structure has been adopted as a core pipeline in numerous subsequent LiDAR odometry studies. However, LiDAR-only approaches exhibit degraded performance under fast dynamic motion and in geometrically degenerate environments such as long corridors or open spaces, where insufficient geometric constraints lead to pose estimation failures.

To handle these challenges, LiDAR-Inertial Odometry (LIO) has gained significant attention by tightly fusing LiDAR measurements with IMU data. LiDAR-IMU fusion methods are broadly categorized into factor graph-based optimization and Kalman filter-based approaches. LIO-Mapping \cite{ye2019tightly} presented a tightly-coupled structure that simultaneously optimizes LiDAR scan matching and IMU preintegration factors through sliding window factor graph optimization. LIO-SAM \cite{shan2020lio} adopted the iSAM2 \cite{kaess2012isam2}-based incremental smoothing framework for efficient factor graph optimization. However, factor graph-based methods incur computational costs that scale with the optimization window size, limiting real-time processing on resource-constrained platforms.

\begin{figure*}[!t]
\centering
\includegraphics[width=\textwidth]{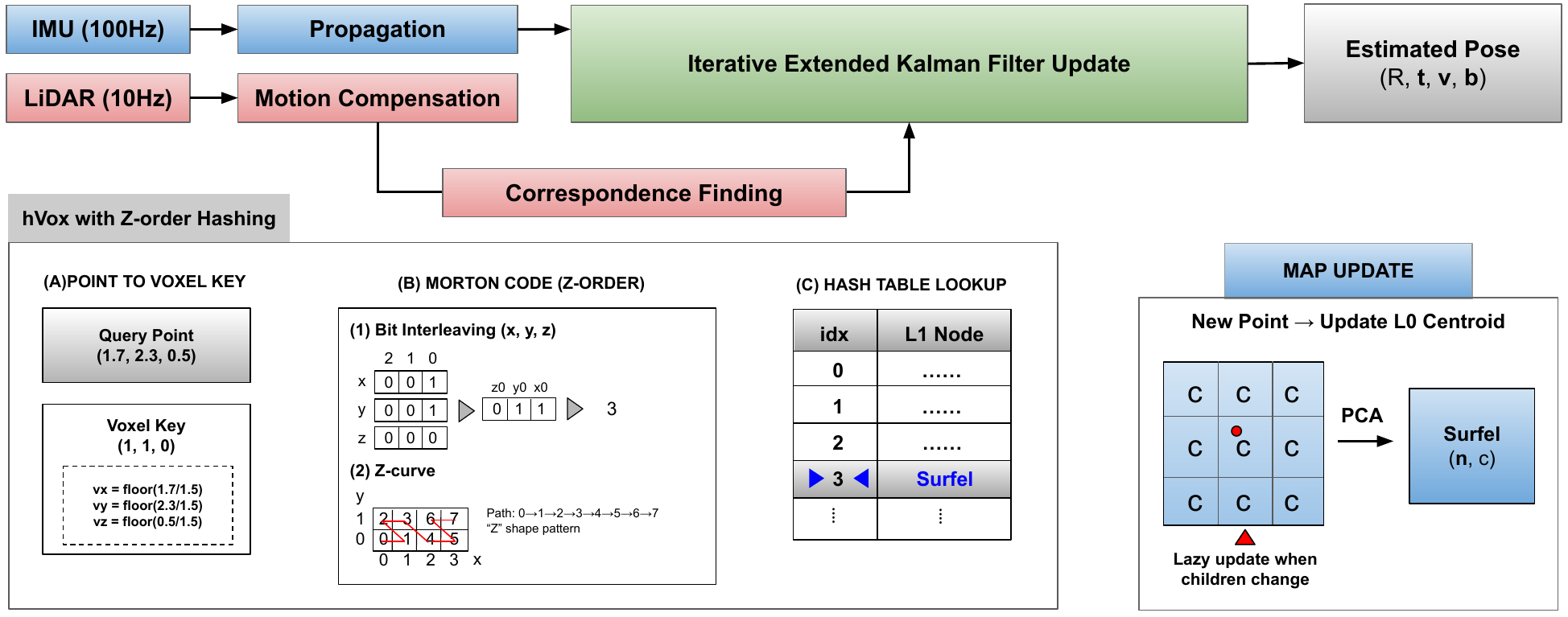}
\caption{Overview of the Surfel-LIO pipeline and correspondence finding procedure. Top: the system fuses IMU propagation and LiDAR measurements through IEKF to estimate pose. Bottom left: correspondence finding via hVox with Z-order hashing, consisting of (A) point-to-voxel key conversion, (B) Morton code computation through bit interleaving, and (C) O(1) hash table lookup to retrieve the pre-computed surfel. Bottom right: map update procedure where new points update L0 centroids, triggering lazy surfel recomputation via PCA when children change. The Morton code example uses 3-bit coordinates for illustration; the actual implementation uses 21 bits per axis to support large-scale environments.}
\label{fig:pipeline}
\end{figure*}

In contrast, Kalman filter-based approaches maintain relatively low computational complexity by performing state propagation and measurement updates sequentially. LINS \cite{qin2020lins} introduced the iterated extended Kalman filter (IEKF) with robocentric formulation for LiDAR-IMU fusion. Fast-LIO \cite{xu2021fast} inherited the IEKF framework from LINS while adopting a direct registration approach that uses raw points without feature extraction, improving computational efficiency. Furthermore, by formulating Kalman gain computation in state dimension rather than measurement dimension, it efficiently processes thousands of points. This IEKF-based structure has become the core estimation framework for subsequent Fast-LIO series research.

Map data structure design is a critical factor determining real-time processing performance in LIO systems. Since nearest neighbor (NN) search and plane estimation must be repeatedly performed for thousands to tens of thousands of points per scan, the complexity of these operations dominates overall processing time. Fast-LIO \cite{xu2021fast} performed NN search by completely rebuilding the k-d tree every frame, incurring $O(N \log N)$ construction cost and inevitable performance degradation as map size increases.

To reduce this overhead, Fast-LIO2 \cite{xu2022fast} proposed the incremental k-d tree (ikd-Tree) \cite{cai2021ikd}. Unlike conventional k-d trees that require costly reconstruction when points are added, the ikd-Tree supports incremental insertion, deletion, and re-balancing, enabling dynamic map updates without tree reconstruction while maintaining $O(\log N)$ complexity for both insertion and NN search. It also efficiently manages local maps during robot movement through box-wise deletion operations. However, computing point-to-plane residuals requires searching for $k$ neighboring points per query point and then performing Principal Component Analysis (PCA) on them to estimate plane parameters. Since covariance computation and eigenvalue decomposition of a $3 \times k$ matrix are required, plane estimation incurs additional $O(k^2)$ complexity. Thus, the total complexity per query is $O(\log N + k^2)$.

Faster-LIO \cite{bai2022faster} proposed spatial hashing-based incremental voxels (iVox) instead of tree structures. iVox partitions 3D space into uniform voxels and stores each voxel in a hash table, enabling $O(1)$ access to any individual voxel. However, since a single voxel often contains insufficient points for reliable plane estimation, iVox must examine multiple neighboring voxels---typically 7 to 27 depending on configuration---to gather enough points. This neighboring voxel search introduces $O(m)$ complexity where $m$ is the number of voxels examined. Furthermore, Faster-LIO still performs PCA at query time for plane estimation, maintaining $O(k^2)$ plane estimation complexity. Consequently, the total complexity per query is $O(m + k^2)$.

In this manuscript, we propose a novel map representation combining hierarchical voxel maps (hVox) with pre-computed surfels. hVox adopts a coarse-to-fine 2-level hierarchical structure (L1$\rightarrow$L0) to partition space hierarchically, with Morton code-based spatial hashing \cite{morton1966computer} at each level enabling $O(1)$ voxel access. Morton codes encode 3D coordinates into a single integer, arranging spatially adjacent voxels adjacently in the hash table to improve cache efficiency.

\begin{figure*}[!t]
\centering
\includegraphics[width=\textwidth]{figures/LIO_VOXEL.pdf}
\caption{Hierarchical voxel structure and surfel representation. (A) An L1 voxel encompasses a $3 \times 3 \times 3$ grid of L0 child voxels, where shaded cells indicate occupied voxels. (B) Each occupied L0 voxel stores only its centroid (red points), computed incrementally as points accumulate. (C) A surfel is derived by applying PCA to the L0 centroids, yielding a compact plane representation $(\mathbf{n}, \mathbf{c})$ for point-to-plane registration.}
\label{fig:hvox_surfel}
\end{figure*}

The key differentiator lies in pre-computing plane information. While existing methods collect neighboring points at query time to perform PCA, our method incrementally updates plane parameters (normal vector, centroid, point count) whenever points are inserted into each voxel. Specifically, when a new point is added to a voxel, the running covariance matrix is updated with $O(1)$ operations using Welford's algorithm \cite{welford1962note}, and once sufficient points accumulate, plane parameters are computed via eigenvalue decomposition and stored as surfels \cite{pfister2000surfels}. Subsequently, at query time, plane information can be immediately obtained by directly referencing the voxel's surfel, eliminating the need for separate PCA computation.

Through this approach, the proposed method achieves $O(1)$ complexity for both NN search and plane estimation. This advantage becomes particularly pronounced during IEKF iterative updates, where plane information for the same points must be referenced multiple times. We provide a comparative summary of existing LIO methods in Table~\ref{tab:lio_comparison}.

\section{Methodology}\label{sec:methodology}

\subsection{System Overview}
\label{sec:overview}

Fig.~\ref{fig:pipeline} presents the overall architecture of Surfel-LIO, a tightly-coupled LiDAR-inertial odometry system designed for real-time state estimation. The system processes IMU measurements at high frequency (100-400Hz) and LiDAR scans at lower frequency (10Hz), fusing them through an Iterated Extended Kalman Filter (IEKF).

Upon receiving an IMU measurement, we propagate the state estimate using standard IMU kinematics, predicting the rotation, position, and velocity of the sensor platform. When a LiDAR scan arrives, the raw point cloud is first undistorted using the IMU-propagated trajectory to compensate for motion during the scan acquisition. Each point is then transformed to the world frame using the current state estimate and queried against our hierarchical voxel map to find its corresponding surfel. Unlike existing methods that perform nearest neighbor search followed by plane fitting, our approach retrieves pre-computed plane parameters directly from the voxel map in O(1) time. The retrieved surfels define point-to-plane residuals that drive the IEKF measurement update, which iterates until convergence. Finally, the optimized pose is used to add the registered scan points to the map, where L0 voxel centroids are updated incrementally and L1 surfels are recomputed as needed.

\subsection{Hierarchical Voxel Map}
\label{sec:hvox}

We propose a two-level hierarchical voxel map (hVox) that separates point aggregation from plane representation, enabling efficient incremental updates while providing accurate geometric information for registration. Fig.~\ref{fig:hvox_surfel} illustrates the structure.

The finest level, Level-0 (L0), consists of small voxels with edge length $s_0$. Rather than storing raw points, each L0 voxel maintains only the incrementally updated centroid of all points that have fallen within it (Fig.~\ref{fig:hvox_surfel}(B)):
\begin{equation}
\mathbf{c}^{(n+1)} = \frac{n \cdot \mathbf{c}^{(n)} + \mathbf{p}_{\text{new}}}{n + 1}
\label{eq:incremental_centroid}
\end{equation}
where $n$ is the accumulated point count and $\mathbf{p}_{\text{new}}$ is the newly inserted point. This formulation requires only O(1) storage per voxel regardless of how many points it contains.

Level-1 (L1) voxels serve as parent nodes, each encompassing a $3 \times 3 \times 3$ grid of L0 children with edge length $s_1 = 3s_0$ (Fig.~\ref{fig:hvox_surfel}(A)). Instead of aggregating points, each L1 voxel stores a pre-computed surfel---a compact planar primitive derived from the centroids of its occupied L0 children (Fig.~\ref{fig:hvox_surfel}(C)). Given the set of centroids $\{\mathbf{c}_i\}_{i=1}^{m}$ from $m$ occupied L0 children, we compute the surfel via Principal Component Analysis\cite{pearson1901pca}. The mean centroid is:
\begin{equation}
\bar{\mathbf{c}} = \frac{1}{m} \sum_{i=1}^{m} \mathbf{c}_i
\label{eq:surfel_centroid}
\end{equation}
The covariance matrix is constructed as:
\begin{equation}
\mathbf{C} = \frac{1}{m} \sum_{i=1}^{m} (\mathbf{c}_i - \bar{\mathbf{c}})(\mathbf{c}_i - \bar{\mathbf{c}})^\top
\label{eq:covariance}
\end{equation}
Eigendecomposition of $\mathbf{C}$ yields eigenvalues $\lambda_1 \geq \lambda_2 \geq \lambda_3$ with corresponding eigenvectors $\mathbf{v}_1, \mathbf{v}_2, \mathbf{v}_3$. The surfel normal is taken as the eigenvector corresponding to the smallest eigenvalue:
\begin{equation}
\mathbf{n} = \mathbf{v}_3
\label{eq:surfel_normal}
\end{equation}
representing the direction of least variance. The planarity score quantifies fitting confidence:
\begin{equation}
\rho = \frac{\lambda_2 - \lambda_3}{\lambda_1 + \epsilon}
\label{eq:planarity}
\end{equation}
where $\epsilon$ is a small constant for numerical stability. Surfels with $\rho$ below a threshold are rejected during correspondence search to avoid unreliable measurements.

To avoid redundant computation, we employ a lazy update strategy: when a new point modifies an L0 centroid, the parent L1 surfel is marked dirty rather than immediately recomputed. This is effective because multiple points from a single scan often fall into the same L1 voxel, and L1 voxels never queried during registration need not be updated at all. Algorithm~\ref{alg:map_update} summarizes the map update procedure.

\begin{algorithm}[t]
\caption{Map Update}
\label{alg:map_update}
\begin{algorithmic}[1]
\Require Point cloud $\mathcal{P}$, transformation $\mathbf{T}_{WL}$
\For{each point $\mathbf{p} \in \mathcal{P}$}
    \State $\mathbf{p}^W \gets \mathbf{T}_{WL} \cdot \mathbf{p}$
    \State $\mathbf{k}_0 \gets \lfloor \mathbf{p}^W / s_0 \rfloor$
    \If{L0[$\mathbf{k}_0$] exists}
        \State Update centroid via Eq.~\ref{eq:incremental_centroid}
    \Else
        \State Create L0[$\mathbf{k}_0$] with centroid $\mathbf{p}^W$
        \State $\mathbf{k}_1 \gets \lfloor \mathbf{k}_0 / 3 \rfloor$
        \State Register L0[$\mathbf{k}_0$] as child of L1[$\mathbf{k}_1$]
    \EndIf
    \State Mark L1[$\mathbf{k}_1$] surfel as dirty
\EndFor
\For{each dirty L1 voxel}
    \State Recompute surfel via Eq.~\ref{eq:surfel_centroid}--\ref{eq:planarity}
\EndFor
\end{algorithmic}
\end{algorithm}

\subsection{Z-order Spatial Hashing}
\label{sec:zorder}

Both L0 and L1 voxels are stored in hash tables for O(1) average-case access. However, standard hash tables suffer from poor cache locality: spatially adjacent voxels may be scattered across memory, causing frequent cache misses as the map grows. Since modern CPUs access RAM in cache lines (typically 64 bytes), loading a single voxel from a random memory location incurs the full memory latency ($\sim$100 cycles), whereas accessing nearby data already in CPU cache costs only $\sim$4 cycles. This disparity makes cache-friendly memory layouts critical for practical performance.

We employ Z-order (Morton) encoding\cite{morton1966computer, samet2006foundations} as our hash function to preserve spatial locality. The bottom-left of Fig.~\ref{fig:pipeline} illustrates the complete correspondence finding procedure using this encoding. Given a query point, step (A) converts the 3D coordinate to an integer voxel key by dividing by the voxel size and flooring. For example, a point $(1.7, 2.3, 0.5)$ with voxel size $s=1.5$ yields key $(1, 1, 0)$. Step (B) computes the Morton code by interleaving the bits of each coordinate in $z, y, x$ order. As shown in Fig.~\ref{fig:pipeline}(B), coordinates $x=1, y=1, z=0$ (binary: 001, 001, 000) produce interleaved bits $z_0y_0x_0 = 011$, yielding Morton code 3. The Z-curve diagram shows how this interleaving traces a Z-shaped path through 2D space (path: $0 \rightarrow 1 \rightarrow 2 \rightarrow 3 \rightarrow \cdots$), naturally clustering spatially adjacent cells into consecutive indices.

The Morton code is computed as:
\begin{equation}
M(k_x, k_y, k_z) = \sum_{i=0}^{20} \left( x_i \cdot 4^i + y_i \cdot 2 \cdot 4^i + z_i \cdot 4 \cdot 4^i \right)
\end{equation}
where $x_i, y_i, z_i$ are the $i$-th bits of each coordinate, and 21 bits per axis support coordinates up to $\pm 2^{20}$ voxels from the origin. This encoding ensures that consecutive LiDAR points---which typically fall in nearby spatial regions---map to similar Morton codes and thus access neighboring hash table entries, improving cache hit rates. Algorithm~\ref{alg:correspondence} summarizes the correspondence finding procedure.

\begin{algorithm}[t]
\caption{Correspondence Finding}
\label{alg:correspondence}
\begin{algorithmic}[1]
\Require Point $\mathbf{p}^W$ in world frame, voxel size $s_1$
\Ensure Surfel $(\mathbf{n}, \mathbf{c}, \rho)$ or $\emptyset$
\State $\mathbf{k} \gets \lfloor \mathbf{p}^W / s_1 \rfloor$ \Comment{Coordinate quantization}
\State $m \gets \textsc{Morton}(k_x, k_y, k_z)$ \Comment{Bit interleaving}
\If{L1[$m$] exists \textbf{and} L1[$m$].$\rho > \rho_{\min}$}
    \State \Return L1[$m$].surfel
\Else
    \State \Return $\emptyset$ \Comment{No valid correspondence}
\EndIf
\end{algorithmic}
\end{algorithm}

Finally, step (C) uses the Morton code as the hash key to perform O(1) lookup into the hash table, directly retrieving the L1 voxel and its pre-computed surfel. This three-step process---coordinate quantization, Morton encoding, and hash lookup---replaces the traditional kNN search and plane fitting with a single hash table access.

For the hash table implementation, we adopt Robin Hood hashing~\cite{celis1986robin}, an open-addressing scheme that minimizes probe sequence variance by redistributing entries based on their displacement from the ideal position. Combined with Morton encoding, this provides both spatial locality in key distribution and memory locality in collision resolution, further improving cache efficiency.

An alternative approach is the Pseudo-Hilbert Curve (PHC) used in Faster-LIO~\cite{bai2022faster}, which provides even better locality preservation than Z-order. However, PHC requires maintaining a sorted data structure and periodic re-sorting as new voxels are inserted, adding computational overhead. In contrast, Morton encoding is computed directly from coordinates via simple bit operations, requiring no auxiliary data structures. Given that our surfel pre-computation already shifts computational burden to the update phase, we opt for the lightweight Morton encoding to keep hashing overhead minimal.

\subsection{State Estimation}
\label{sec:iekf}

We employ an Iterated Extended Kalman Filter (IEKF) to fuse IMU measurements with LiDAR observations, following the error-state formulation commonly used in LIO systems~\cite{xu2021fast}. The state vector is defined as:
\begin{equation}
\mathbf{x} = \begin{bmatrix} \mathbf{R}^\top & \mathbf{p}^\top & \mathbf{v}^\top & \mathbf{b}_g^\top & \mathbf{b}_a^\top \end{bmatrix}^\top \in SO(3) \times \mathbb{R}^{12}
\label{eq:state}
\end{equation}
where $\mathbf{R} \in SO(3)$ is the rotation from body to world frame, $\mathbf{p} \in \mathbb{R}^3$ is the position, $\mathbf{v} \in \mathbb{R}^3$ is the velocity, and $\mathbf{b}_g, \mathbf{b}_a \in \mathbb{R}^3$ are the gyroscope and accelerometer biases respectively.

Upon receiving IMU measurements, the state is propagated using the discrete-time kinematics:
\begin{equation}
\hat{\mathbf{R}}_{k+1} = \hat{\mathbf{R}}_k \cdot \text{Exp}((\boldsymbol{\omega}_k - \mathbf{b}_g) \Delta t)
\label{eq:imu_rot}
\end{equation}
\begin{equation}
\hat{\mathbf{v}}_{k+1} = \hat{\mathbf{v}}_k + (\hat{\mathbf{R}}_k (\mathbf{a}_k - \mathbf{b}_a) + \mathbf{g}) \Delta t
\label{eq:imu_vel}
\end{equation}
\begin{equation}
\hat{\mathbf{p}}_{k+1} = \hat{\mathbf{p}}_k + \hat{\mathbf{v}}_k \Delta t + \frac{1}{2}(\hat{\mathbf{R}}_k (\mathbf{a}_k - \mathbf{b}_a) + \mathbf{g}) \Delta t^2
\label{eq:imu_pos}
\end{equation}
where $\boldsymbol{\omega}_k$ and $\mathbf{a}_k$ are the gyroscope and accelerometer readings, $\mathbf{g}$ is the gravity vector, $\Delta t$ is the time interval, and $\text{Exp}(\cdot)$ denotes the exponential map from $\mathfrak{so}(3)$ to $SO(3)$.

When a LiDAR scan arrives, each point $\mathbf{p}_i^L$ in the LiDAR frame is transformed to the world frame and matched to its corresponding surfel via the hash table lookup described in Section~\ref{sec:zorder}. Given a surfel with normal $\mathbf{n}_i$ and centroid $\mathbf{c}_i$, the point-to-plane residual is:
\begin{equation}
r_i = \mathbf{n}_i^\top \left( \mathbf{R} \mathbf{p}_i^L + \mathbf{p} - \mathbf{c}_i \right)
\label{eq:residual}
\end{equation}
The IEKF iteratively refines the state estimate by linearizing this residual around the current estimate. Let $\delta \mathbf{x} = [\delta \boldsymbol{\theta}^\top, \delta \mathbf{p}^\top, \delta \mathbf{v}^\top, \delta \mathbf{b}_g^\top, \delta \mathbf{b}_a^\top]^\top$ denote the error state, where $\delta \boldsymbol{\theta} \in \mathbb{R}^3$ parameterizes the rotation error via $\mathbf{R}_{\text{true}} = \mathbf{R} \cdot \text{Exp}(\delta \boldsymbol{\theta})$. The Jacobian of the residual with respect to the error state is:
\begin{equation}
\mathbf{H}_i = \frac{\partial r_i}{\partial \delta \mathbf{x}} = \begin{bmatrix} -\mathbf{n}_i^\top \mathbf{R} [\mathbf{p}_i^L]_\times & \mathbf{n}_i^\top & \mathbf{0}_{1 \times 9} \end{bmatrix}
\label{eq:jacobian}
\end{equation}
where $[\cdot]_\times$ denotes the skew-symmetric matrix. Stacking all valid correspondences, the measurement update follows the standard Kalman filter equations with iteration until the state correction converges below a threshold.

\begin{table}[t]
\centering
\caption{Per-operation timing comparison averaged over all sequences.}
\label{tab:timing}
\renewcommand{\arraystretch}{1.3}
\begin{tabularx}{\columnwidth}{l*{3}{>{\centering\arraybackslash}X}}
\toprule
\toprule
\textbf{Operation} & \textbf{Fast-LIO2} & \textbf{Faster-LIO} & \textbf{Surfel-LIO} \\
\midrule
NN Search ($\mu s$/pt) & 1.42 & 2.76 & \textbf{0.05} \\
Plane Est. ($\mu s$/pt) & 0.17 & 0.61 & \textbf{0.01} \\
Map Update ($ms$/fr) & 0.34 & \textbf{0.03} & \textbf{0.03} \\
\bottomrule
\bottomrule
\end{tabularx}
\end{table}

\begin{table*}[t]
\centering
\caption{Detailed results on M3DGR dataset. APE denotes Absolute Pose Error RMSE ($m$), FPS denotes frames per second. \textcolor{red}{\textbf{Red bold}} indicates best, \textcolor{blue}{\textbf{blue bold}} indicates second best.}
\label{tab:detailed_results}
\renewcommand{\arraystretch}{1.2}
\setlength{\tabcolsep}{4pt}
\begin{tabularx}{\textwidth}{l*{12}{>{\centering\arraybackslash}X}}
\toprule
\toprule
 & \multicolumn{6}{c}{\textbf{Livox AVIA}} & \multicolumn{6}{c}{\textbf{Livox Mid360}} \\
\cmidrule(lr){2-7} \cmidrule(lr){8-13}
 & \multicolumn{2}{c}{\textbf{Fast-LIO2}} & \multicolumn{2}{c}{\textbf{Faster-LIO}} & \multicolumn{2}{c}{\textbf{Surfel-LIO}} & \multicolumn{2}{c}{\textbf{Fast-LIO2}} & \multicolumn{2}{c}{\textbf{Faster-LIO}} & \multicolumn{2}{c}{\textbf{Surfel-LIO}} \\
\cmidrule(lr){2-3} \cmidrule(lr){4-5} \cmidrule(lr){6-7} \cmidrule(lr){8-9} \cmidrule(lr){10-11} \cmidrule(lr){12-13}
Sequence & APE & FPS & APE & FPS & APE & FPS & APE & FPS & APE & FPS & APE & FPS \\
\midrule
Dark01 & 0.258 & 140 & 0.223 & 203 & 0.118 & 670 & 0.177 & 589 & 0.173 & 925 & 0.185 & 1044 \\
Dark02 & 0.729 & 120 & 0.645 & 195 & 0.692 & 488 & 0.239 & 337 & 0.212 & 401 & 0.310 & 720 \\
Dynamic03 & 0.165 & 145 & 0.151 & 225 & 0.266 & 606 & 0.178 & 263 & 0.178 & 461 & 0.206 & 670 \\
Dynamic04 & 0.279 & 138 & 0.261 & 201 & 0.392 & 603 & 0.216 & 265 & 0.214 & 363 & 0.246 & 657 \\
Occlusion03 & 0.257 & 124 & 0.283 & 179 & 0.271 & 561 & 0.423 & 301 & 0.463 & 373 & 0.315 & 687 \\
Occlusion04 & 0.479 & 130 & 0.337 & 203 & 0.295 & 501 & 0.216 & 251 & 0.284 & 385 & 0.345 & 596 \\
Varying-illu03 & 0.897 & 120 & 1.032 & 165 & 0.961 & 436 & 1.221 & 242 & 1.189 & 259 & 0.957 & 618 \\
Varying-illu04 & 0.102 & 93 & 0.119 & 144 & 0.125 & 435 & 0.161 & 205 & 0.163 & 170 & 0.206 & 664 \\
Varying-illu05 & 0.402 & 133 & 0.207 & 172 & 0.167 & 576 & 0.245 & 297 & 0.290 & 498 & 0.307 & 704 \\
\midrule
Average & 0.397 & 125 & \textcolor{red}{\textbf{0.362}} & \textcolor{blue}{\textbf{184}} & \textcolor{blue}{\textbf{0.365}} & \textcolor{red}{\textbf{531}} & \textcolor{red}{\textbf{0.342}} & 282 & \textcolor{blue}{\textbf{0.352}} & \textcolor{blue}{\textbf{353}} & \textcolor{red}{\textbf{0.342}} & \textcolor{red}{\textbf{690}} \\
\bottomrule
\bottomrule
\end{tabularx}
\end{table*}

\section{Experimental Result}\label{sec:experiments}

\subsection{Experimental Setup}

We evaluate our method on the M3DGR dataset~\cite{zhang2025m3dgr}, a multi-sensor dataset designed for ground robot localization. We select nine sequences (Dark01-02, Dynamic03-04, Occlusion03-04, Varying-illu03-05) that provide centimeter-level ground truth from motion capture systems and RTK-GPS. These sequences present various challenges: Dark sequences feature low-light indoor environments where visual sensors struggle, Dynamic sequences contain moving pedestrians and objects that may corrupt map consistency, Occlusion sequences include partial sensor blockage from nearby obstacles, and Varying-illu sequences exhibit rapid lighting changes during indoor-outdoor transitions.

The dataset provides two Livox LiDAR sensors with different characteristics: AVIA with non-repetitive scanning pattern and 70° circular FOV generating approximately 24,000 points per scan, and Mid-360 with 360° horizontal FOV generating approximately 15,000 points per scan. This setup allows us to examine how different point densities and scanning patterns affect each method's computational load and estimation quality.

We compare against Fast-LIO2~\cite{xu2022fast} and Faster-LIO~\cite{bai2022faster}, two widely-used tightly-coupled LIO systems representing different approaches to map management. Fast-LIO2 employs an incremental k-d tree (ikd-Tree) that supports efficient point insertion and deletion with $O(\log N)$ search complexity, where $N$ denotes the number of map points. Faster-LIO uses hash-based incremental voxels (iVox) that provides $O(1)$ voxel access, but gathering sufficient neighbors for plane fitting requires examining 7--27 surrounding voxels depending on the query point's position relative to voxel boundaries.

All experiments are conducted on a desktop PC with Intel Core i7-14700KF CPU (20 cores, 5.6GHz boost) and 64GB DDR5 RAM using official open-source implementations. To ensure fair comparison, all methods use identical voxel resolution (0.5$m$) and local map extent (200$m$ $\times$ 200$m$).

\subsection{Per-Operation Timing Analysis}

To understand the computational characteristics of different map representations, we instrument each method to measure per-point execution time for core operations: nearest neighbor search, plane estimation, and map update. Table~\ref{tab:timing} presents the timing breakdown averaged over all sequences.

For nearest neighbor search, Fast-LIO2 requires 1.42$\mu s$ per point due to ikd-Tree traversal, where search time grows logarithmically with the number of map points. Faster-LIO requires 2.76$\mu s$ per point despite using hash-based voxel indexing; this is because finding $k$ nearest neighbors for plane fitting requires examining multiple neighboring voxels, and each voxel may contain multiple points that must be distance-sorted. Our method requires only 0.05$\mu s$ per point, as the query reduces to a single hash table lookup to retrieve the voxel containing the query point, which directly provides pre-computed plane parameters without additional neighbor gathering.

For plane estimation, Fast-LIO2 and Faster-LIO compute plane parameters at query time using SVD-based fitting on retrieved neighbor points, requiring 0.17$\mu s$ and 0.61$\mu s$ per point respectively. The difference may be attributed to implementation details and the number of points used for fitting. Our method requires only 0.01$\mu s$ per point since plane parameters (normal vector and centroid) are pre-computed and stored as surfels during map update; the residual calculation reduces to a simple dot product between the query point and the stored normal vector.

For map updates, Fast-LIO2 requires approximately 0.34$ms$ per frame due to the costs associated with incremental tree balancing. In contrast, both Faster-LIO and our method demonstrate superior efficiency, averaging around 0.03$ms$. While Faster-LIO achieves this by updating hash table entries, our method updates surfel statistics through incremental covariance computation. This comparable performance with the hash-based approach suggests that the overhead of maintaining pre-computed surfels is negligible.

\subsection{Odometry Accuracy and Efficiency}

Table~\ref{tab:detailed_results} presents end-to-end performance for both LiDAR sensors. APE (Absolute Pose Error) is computed as the root mean square error of translational differences between estimated and ground truth poses after SE(3) alignment using the evo evaluation toolkit~\cite{grupp2017evo}.

On AVIA sequences with higher point density ($\sim$24,000 points/scan), Surfel-LIO processes at 531 FPS on average, compared to 125 FPS for Fast-LIO2 and 184 FPS for Faster-LIO. On Mid-360 sequences with lower point density ($\sim$15,000 points/scan), all methods achieve higher throughput: Surfel-LIO at 690 FPS, Fast-LIO2 at 282 FPS, and Faster-LIO at 353 FPS. The speed improvement is more pronounced on denser point clouds, which is consistent with the expectation that per-point query savings accumulate with more points per frame.

Regarding state estimation accuracy, all three methods achieve comparable performance across the tested sequences. On AVIA sequences, average APE is 0.397$m$ for Fast-LIO2, 0.362$m$ for Faster-LIO, and 0.365$m$ for Surfel-LIO. On Mid-360 sequences, the three methods show similar accuracy: 0.342$m$ for Fast-LIO2, 0.352$m$ for Faster-LIO, and 0.342$m$ for Surfel-LIO. Challenging sequences such as Dark02 and Varying-illu03 show higher errors across all methods, likely due to aggressive motion during low-light conditions and rapid illumination changes.

These results suggest that the surfel-based map representation can improve processing efficiency while maintaining state estimation accuracy comparable to existing point-based methods. The consistent accuracy across methods indicates that the point-to-plane residual formulation is robust to different underlying map structures, with the primary distinction being how efficiently each method computes these residuals during iterative state estimation.

\section{Conclusion}\label{sec:conclusion}

We presented Surfel-LIO, a LiDAR-inertial odometry system that employs pre-computed surfels for efficient correspondence search. By storing plane parameters as surfels during map construction, our approach reduces per-point query complexity from $O(\log N)$ or $O(m)$ to $O(1)$, eliminating the need for runtime nearest neighbor search and plane fitting.

Experimental results on the M3DGR dataset suggest that this design can improve processing efficiency compared to existing methods. Surfel-LIO achieved higher frame rates (531 FPS on AVIA, 690 FPS on Mid-360) while maintaining state estimation accuracy comparable to Fast-LIO2 and Faster-LIO. The per-operation timing analysis indicates that the computational savings primarily come from reduced query overhead, while the map update cost remains similar across all methods.

However, our approach has limitations that warrant further investigation. First, the surfel representation assumes locally planar geometry, which may be less suitable for environments with complex curved surfaces or sparse structures. Second, the current implementation uses a fixed voxel resolution, whereas adaptive resolution strategies could potentially improve both accuracy and efficiency. Third, our evaluation is limited to ground robot scenarios with Livox LiDAR sensors; the generalization to other platforms (e.g., aerial robots, handheld devices) and sensor types (e.g., spinning LiDARs) remains to be validated.

Future work includes extending the framework to handle non-planar geometric primitives such as lines and corners, investigating adaptive voxel sizing based on local geometry complexity, and integrating loop closure for globally consistent mapping. We also plan to explore the applicability of the surfel-based representation to other state estimation problems beyond LiDAR-inertial odometry.

\bibliography{references}

@article{cadena2016past,
  title={Past, present, and future of simultaneous localization and mapping: Toward the robust-perception age},
  author={Cadena, Cesar and Carlone, Luca and Carrillo, Henry and Latif, Yasir and Scaramuzza, Davide and Neira, Jos{\'e} and Reid, Ian and Leonard, John J},
  journal={IEEE Transactions on Robotics},
  volume={32},
  number={6},
  pages={1309--1332},
  year={2016}
}

@article{placed2023survey,
  title={A Survey on Active Simultaneous Localization and Mapping: State of the Art and New Frontiers},
  author={Placed, Julio A and Strader, Jared and Carrillo, Henry and Atanasov, Nikolay and Indelman, Vadim and Carlone, Luca and Castellanos, Jos{\'e} A},
  journal={IEEE Transactions on Robotics},
  year={2023}
}

@inproceedings{shan2020lio,
  title={LIO-SAM: Tightly-coupled Lidar Inertial Odometry via Smoothing and Mapping},
  author={Shan, Tixiao and Englot, Brendan and Meyers, Drew and Wang, Wei and Ratti, Carlo and Rus, Daniela},
  booktitle={IEEE/RSJ International Conference on Intelligent Robots and Systems (IROS)},
  pages={5135--5142},
  year={2020}
}

@inproceedings{qin2020lins,
  title={LINS: A Lidar-Inertial State Estimator for Robust and Efficient Navigation},
  author={Qin, Chao and Ye, Haoyang and Pranata, Christian E and Han, Jun and Zhang, Shuyang and Liu, Ming},
  booktitle={IEEE International Conference on Robotics and Automation (ICRA)},
  pages={8899--8906},
  year={2020}
}

@article{xu2022fast,
  title={FAST-LIO2: Fast Direct LiDAR-Inertial Odometry},
  author={Xu, Wei and Cai, Yixi and He, Dongjiao and Lin, Jiarong and Zhang, Fu},
  journal={IEEE Transactions on Robotics},
  volume={38},
  number={4},
  pages={2053--2073},
  year={2022}
}

@article{cai2021ikd,
  title={ikd-Tree: An Incremental KD Tree for Robotic Applications},
  author={Cai, Yixi and Xu, Wei and Zhang, Fu},
  journal={IEEE Robotics and Automation Letters},
  volume={6},
  number={2},
  pages={3564--3571},
  year={2021}
}

@inproceedings{bai2022faster,
  title={FASTER-LIO: Lightweight Tightly Coupled Lidar-Inertial Odometry Using Parallel Sparse Incremental Voxels},
  author={Bai, Chunge and Xiao, Tao and Chen, Yajie and Wang, Haoqian and Zhang, Fang and Gao, Xiang},
  booktitle={IEEE/RSJ International Conference on Intelligent Robots and Systems (IROS)},
  pages={4095--4102},
  year={2022}
}

@article{besl1992method,
  title={A Method for Registration of 3-D Shapes},
  author={Besl, Paul J and McKay, Neil D},
  journal={IEEE Transactions on Pattern Analysis and Machine Intelligence},
  volume={14},
  number={2},
  pages={239--256},
  year={1992}
}

@inproceedings{chen1992object,
  title={Object Modelling by Registration of Multiple Range Images},
  author={Chen, Yang and Medioni, G{\'e}rard},
  booktitle={IEEE International Conference on Robotics and Automation},
  pages={2724--2729},
  year={1992}
}

@article{bell1993iterated,
  title={The Iterated Kalman Filter Update as a Gauss-Newton Method},
  author={Bell, Bradley M and Cathey, Frederick W},
  journal={IEEE Transactions on Automatic Control},
  volume={38},
  number={2},
  pages={294--297},
  year={1993}
}

@article{pearson1901pca,
  title={On Lines and Planes of Closest Fit to Systems of Points in Space},
  author={Pearson, Karl},
  journal={The London, Edinburgh, and Dublin Philosophical Magazine and Journal of Science},
  volume={2},
  number={11},
  pages={559--572},
  year={1901}
}

@techreport{low2004linear,
  title={Linear Least-Squares Optimization for Point-to-Plane ICP Surface Registration},
  author={Low, Kok-Lim},
  institution={University of North Carolina at Chapel Hill},
  year={2004}
}

@article{niessner2013real,
  title={Real-time 3D Reconstruction at Scale Using Voxel Hashing},
  author={Nie{\ss}ner, Matthias and Zollh{\"o}fer, Michael and Izadi, Shahram and Stamminger, Marc},
  journal={ACM Transactions on Graphics (TOG)},
  volume={32},
  number={6},
  pages={1--11},
  year={2013}
}

@inproceedings{teschner2003optimized,
  title={Optimized Spatial Hashing for Collision Detection of Deformable Objects},
  author={Teschner, Matthias and Heidelberger, Bruno and M{\"u}ller, Matthias and Pomerantes, Danat and Gross, Markus H},
  booktitle={VMV},
  volume={3},
  pages={47--54},
  year={2003}
}

@article{hornung2013octomap,
  title={OctoMap: An Efficient Probabilistic 3D Mapping Framework Based on Octrees},
  author={Hornung, Armin and Wurm, Kai M and Bennewitz, Maren and Stachniss, Cyrill and Burgard, Wolfram},
  journal={Autonomous Robots},
  volume={34},
  number={3},
  pages={189--206},
  year={2013}
}

@article{vespa2018efficient,
  title={Efficient Octree-Based Volumetric SLAM Supporting Signed-Distance and Occupancy Mapping},
  author={Vespa, Emanuele and Nikolov, Nikolay and Grimm, Marius and Nardi, Luigi and Kelly, Paul HJ and Leutenegger, Stefan},
  journal={IEEE Robotics and Automation Letters},
  volume={3},
  number={2},
  pages={1144--1151},
  year={2018}
}

@inproceedings{pfister2000surfels,
  title={Surfels: Surface Elements as Rendering Primitives},
  author={Pfister, Hanspeter and Zwicker, Matthias and Van Baar, Jeroen and Gross, Markus},
  booktitle={Proceedings of the 27th Annual Conference on Computer Graphics and Interactive Techniques},
  pages={335--342},
  year={2000}
}

@techreport{morton1966computer,
  title={A Computer Oriented Geodetic Data Base and a New Technique in File Sequencing},
  author={Morton, Guy M},
  institution={IBM Ltd.},
  year={1966}
}

@book{samet2006foundations,
  title={Foundations of Multidimensional and Metric Data Structures},
  author={Samet, Hanan},
  year={2006},
  publisher={Morgan Kaufmann}
}

@article{welford1962note,
  title={Note on a Method for Calculating Corrected Sums of Squares and Products},
  author={Welford, BP},
  journal={Technometrics},
  volume={4},
  number={3},
  pages={419--420},
  year={1962}
}

@article{zhang2025m3dgr,
  title={Towards Robust Sensor-Fusion Ground SLAM: A Comprehensive Benchmark and A Resilient Framework},
  author={Zhang, Deteng and Zhang, Junjie and Sun, Yan and Li, Tao and Yin, Hao and Xie, Hongzhao and Yin, Jie},
  journal={arXiv preprint arXiv:2507.08364},
  year={2025}
}

@inproceedings{zhang2014loam,
  title={{LOAM}: Lidar Odometry and Mapping in Real-time},
  author={Zhang, Ji and Singh, Sanjiv},
  booktitle={Robotics: Science and Systems (RSS)},
  year={2014}
}

@article{xu2021fast,
  title={{FAST-LIO}: A Fast, Robust LiDAR-Inertial Odometry Package by Tightly-Coupled Iterated Kalman Filter},
  author={Xu, Wei and Zhang, Fu},
  journal={IEEE Robotics and Automation Letters},
  volume={6},
  number={2},
  pages={3317--3324},
  year={2021}
}

@inproceedings{ye2019tightly,
  title={Tightly Coupled 3D Lidar Inertial Odometry and Mapping},
  author={Ye, Haoyang and Chen, Yuying and Liu, Ming},
  booktitle={IEEE International Conference on Robotics and Automation (ICRA)},
  pages={3144--3150},
  year={2019}
}

@article{kaess2012isam2,
  title={{iSAM2}: Incremental Smoothing and Mapping Using the Bayes Tree},
  author={Kaess, Michael and Johannsson, Hordur and Roberts, Richard and Ila, Viorela and Leonard, John J and Dellaert, Frank},
  journal={The International Journal of Robotics Research},
  volume={31},
  number={2},
  pages={216--235},
  year={2012}
}

@misc{grupp2017evo,
  author = {Grupp, Michael},
  title = {evo: Python package for the evaluation of odometry and SLAM},
  year = {2017},
  publisher = {GitHub},
  journal = {GitHub repository},
  howpublished = {\url{https://github.com/MichaelGrupp/evo}}
}

@techreport{celis1986robin,
  title={Robin Hood Hashing},
  author={Celis, Pedro and Larson, Per-{\AA}ke and Munro, J. Ian},
  year={1986},
  institution={University of Waterloo},
  number={CS-86-14}
}

\end{document}